\documentclass{article}
\usepackage{ijcai19}

\usepackage{times}
\usepackage{soul}
\usepackage{url}
\usepackage[hidelinks]{hyperref}
\usepackage[utf8]{inputenc}
\usepackage[small]{caption}
\usepackage{graphicx}
\usepackage{amsmath}
\usepackage{booktabs}
\usepackage{algorithm}
\usepackage{algorithmic}
\usepackage{subcaption}
\usepackage{float}
\usepackage{appendix}
\usepackage{tcolorbox}
\usepackage{xcolor} 
\definecolor{darkgreen}{rgb}{0.0, 0.5, 0.0} 
\definecolor{darkyellow}{rgb}{0.6, 0.3, 0.0} 

\title{Natural Language Fine-Tuning}

\author{
Jia Liu$^{1,2}$\and
Yue Wang$^{1,3}$\and
Zhiqi Lin$^{1,3}$\and
Min Chen$^{1,3}$\footnote{Corresponding authors: Min Chen, Yixue Hao, Long Hu}\and
Yixue Hao$^2$\footnotemark[1]\and
Long Hu$^2$\footnotemark[1]\\
\affiliations
$^1$Pazhou Laboratory, Guangzhou, China\\
$^2$Huazhong University of Science and Technology, Wuhan, China\\
$^3$South China University of Technology, Guangzhou, China\\
\emails
\{liujia0330, yixuehao, hulong\}@hust.edu.cn,\\
\{csyuewang, 202311089192\}@mail.scut.edu.cn,\\
minchen@ieee.org
}

\begin{document}

\maketitle

\begin{abstract}
Large language model fine-tuning techniques typically depend on extensive labeled data, external guidance, and feedback, such as human alignment, scalar rewards, and demonstration. However, in practical application, the scarcity of specific knowledge poses unprecedented challenges to existing fine-tuning techniques. In this paper, focusing on fine-tuning tasks in specific domains with limited data, we introduce Natural Language Fine-Tuning (NLFT), which utilizes natural language for fine-tuning for the first time. By leveraging the strong language comprehension capability of the target LM, NLFT attaches the guidance of natural language to the token-level outputs. Then, saliency tokens are identified with calculated probabilities. Since linguistic information is effectively utilized in NLFT, our proposed method significantly reduces training costs. It markedly enhances training efficiency, comprehensively outperforming reinforcement fine-tuning algorithms in accuracy, time-saving, and resource conservation. Additionally, on the macro level, NLFT can be viewed as a token-level fine-grained optimization of SFT, thereby efficiently replacing the SFT process without the need for warm-up (as opposed to ReFT requiring multiple rounds of warm-up with SFT). Compared to SFT, NLFT does not increase the algorithmic complexity, maintaining $O(n)$. Extensive experiments on the GSM8K dataset demonstrate that NLFT, with only 50 data instances, achieves an accuracy increase that exceeds SFT by 219\%. Compared to ReFT, the time complexity and space complexity of NLFT are reduced by 78.27\% and 92.24\%, respectively. The superior technique of NLFT is paving the way for the deployment of various innovative LLM fine-tuning applications when resources are limited at network edges. 

Our code has been released at \url{https://github.com/Julia-LiuJ/NLFT}.
\end{abstract}

\section{Introduction}
\begin{figure}[h]
    \centering
    \includegraphics[width=1\linewidth]{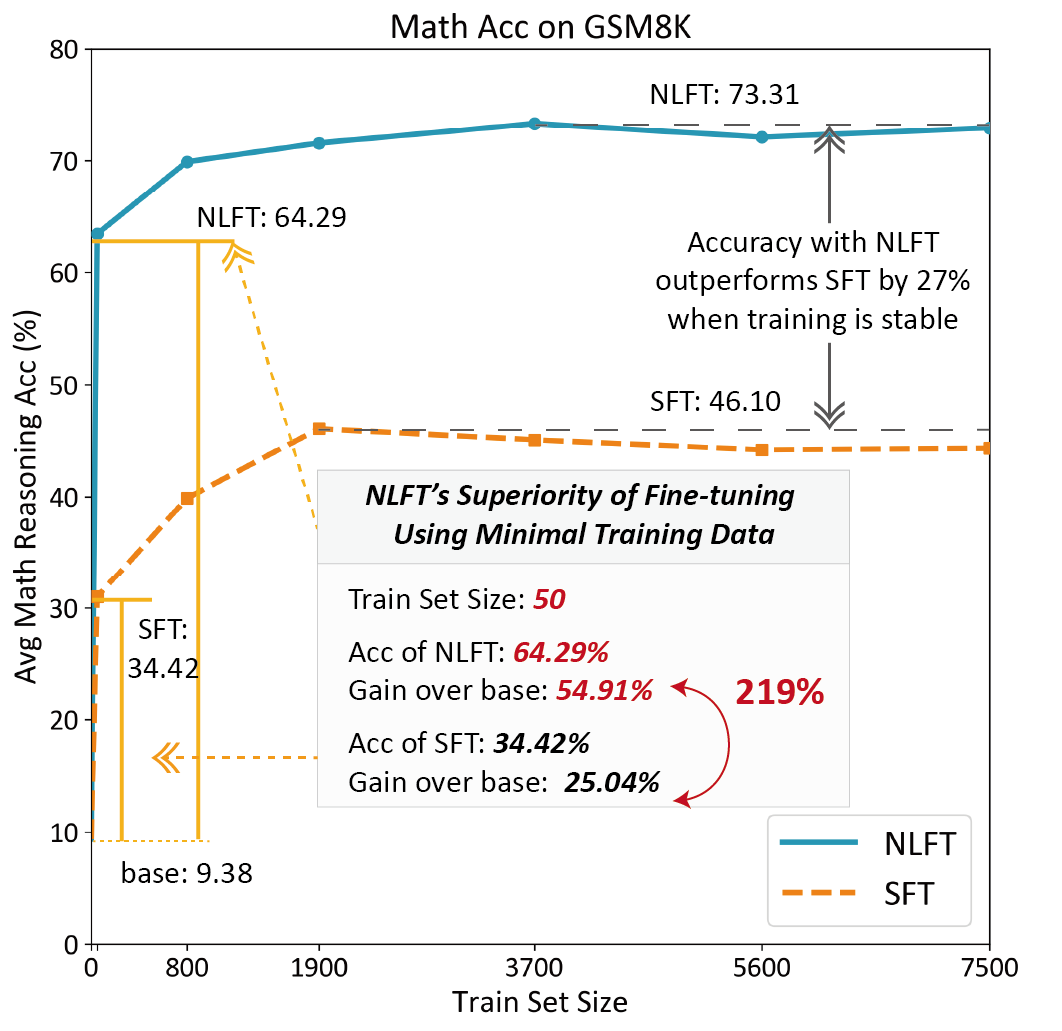}
    \caption{Accuracy Comparison of SFT and NLFT on GSM8K dataset. NLFT has the same time and space complexity as SFT but achieves a 27\% increase in fine-tuning performance, maintaining a stable performance advantage thereafter. With minimal dataset samples (only 50 data points), a brief training period (3 epochs, 287 seconds), and low computational resource consumption (44.46 GB of GPU memory usage), NLFT does not require a warm-up phase and can achieve a performance 1.19 times greater than SFT. According to the ReFT paper, ReFT is unable to outperform SFT within 8 epochs.}
    \label{fig:fig01}
\end{figure}


\begin{figure*}[h]
    \centering
    \includegraphics[width=1\linewidth]{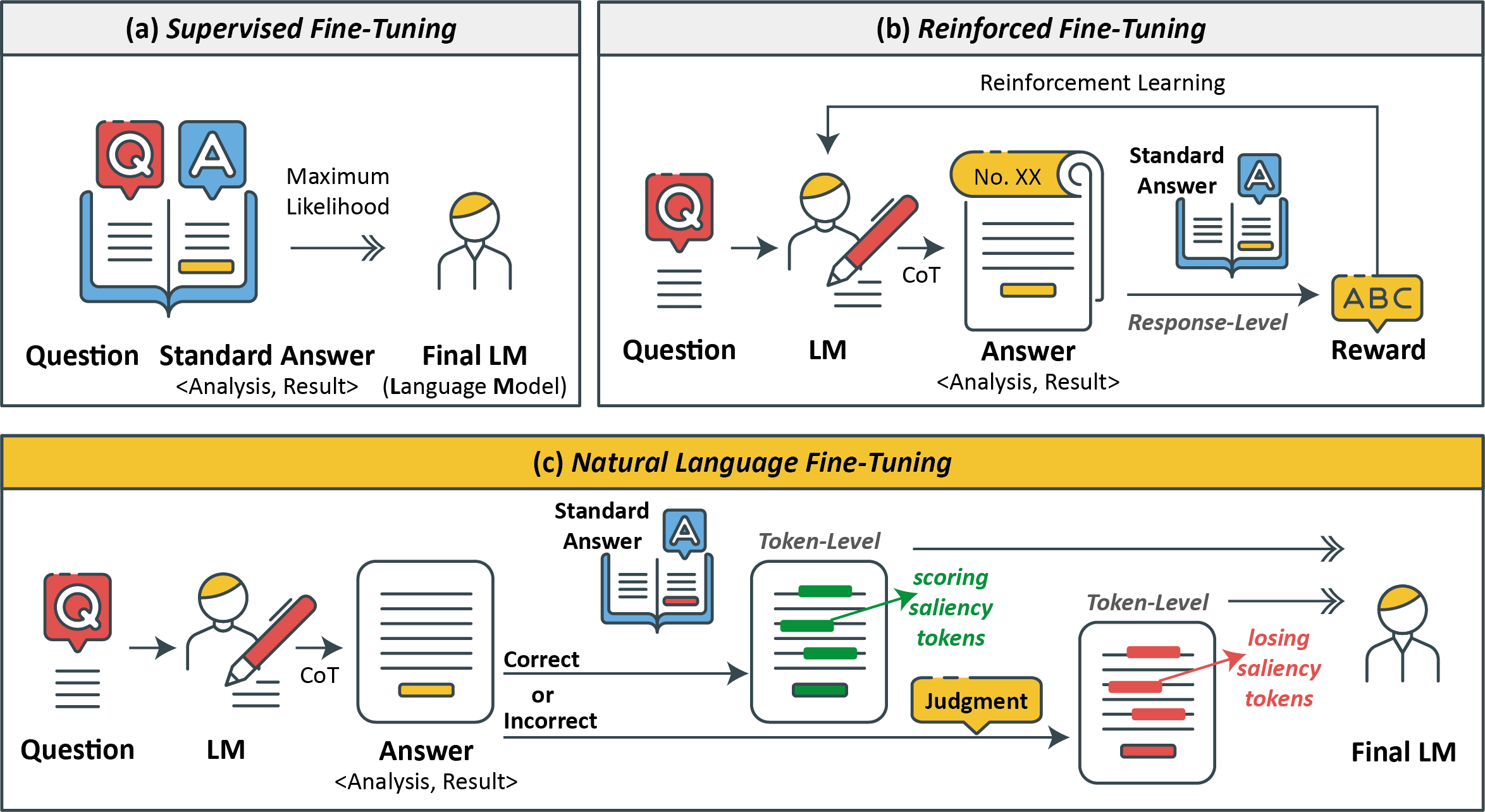}
    \caption{Training process of SFT, ReFT, and NLFT. (a) The training process of SFT, which can be analogous to a student directly learning from a collection of exercises,  which include problems and their reference solutions; (b) The training process of ReFT, which can be realized as a student repeatedly submitting their exam answers to a grading system, receiving scores, and striving to discover the strategies that maximize their marks; (c) The training process of our proposed algorithm NLFT, which is similar to a student submitting exam answers and receiving detailed feedback from a natural language evaluator. The system highlights the scoring points and losing points, allowing the student to learn from both well-graded examples (i.e., for learning from teaching) and their work (i.e., for self-study), thereby increasing their performance.}
    \label{fig:SFT&ReFT&NLFT}
\end{figure*}

Supervised Fine-Tuning (SFT) is the most commonly employed method for fine-tuning Large Language Models (LLMs). As a foundational step in model fine-tuning, its application enables LLMs to better adapt to tasks across various domains and more effectively address specialized issues. For instance, in the alignment tasks, both  Reinforcement Learning with Human Feedback (RLHF)~\cite{ouyang2022training} and Direct Preference Optimization (DPO)~\cite{rafailov2024direct} utilize SFT to provide a solid initial condition for the LLM, thereby leading the learning process to be more efficient and stable. Typically, for mathematical problem solving, researchers employ Chain-of-Thought (CoT)~\cite{wei2022chain} to annotate problems and answers, then use SFT to fine-tune the model for better tackling mathematical questions~\cite{feng2024towards}~\cite{chu2023survey}~\cite{wang2022self}.




As the application field of LLMs continues to expand, fine-tuning with small-scale, domain-specific data remains challenging due to the inefficiency of existing methods in utilizing limited samples. Consequently, researchers are progressively exploring renovated fine-tuning methods to optimize models. Recently, Reinforced Fine-Tuning (ReFT~\cite{luong2024reft} or RFT~\cite{openaiRFT}) has garnered widespread attention in the academic community for its superior performance in terms of accuracy increment. It employs Reinforcement Learning (RL) for the fine-tuning of model parameters, and thus achieving more efficient model optimization within the same data scale. However, introducing RL into LLMs results in a significant increase in time and space complexity, thereby raising the barriers to the deployment and utilization of this technology, especially in a mobile and/or dynamic network environment. Meanwhile, ReFT does not entirely replace SFT. In practical applications, it still requires the SFT to warming-up to ensure that the model can more effectively adapt to specific tasks or datasets~\cite{luong2024reft}



To address these issues, in this paper, we propose a novel minimal data fine-tuning method named Natural Language Fine-Tuning (NLFT). Compared to previous methods, NLFT utilizes natural language as the supervising signal and employs a simple minimal data fine-tuning algorithm to enhance the efficiency of fine-tuning. To better illustrate the differences among SFT, ReFT, and NLFT, we give the following analogy. 





LLM is analogous to a student, and LLM's fine-tuning process is similar to the learning process of the student. Then, SFT, ReFT, and NLFT represent three individual learning processes of the student. Given learning math reasoning as an example, in SFT, the student studies in parrot-fashion, where the student is expected to write down a predetermined answer when seeing some particular question after screening numerous pairs of questions and standard answers. In ReFT, the student first obtains the basic technique of solving math reasoning problems by several epochs of SFT. Then, in order to further improve the technique, ReFT requires the student to submit answer sheets which include the detailed analysis for leading the math problem solution. In ReFT, a score is given for each answer sheet by comparing it to the standard answer. By the score, the student adjusts the strategy of math reasoning, which is similar to reinforcement learning. Since the target of the student is to achieve a high score as much as possible, lots of rounds of submitting the answer sheet and obtaining feedback from the evaluating system are needed. However, in NLFT, the student's learning process more closely resembles a self-study method. When the answer sheet hits majority scoring points, which represents the student is on a good studying track at the beginning, the student will re-answer the exam based on the standard answer and compare it with the previous one to identify the scoring points (see the upper part in Fig. 2c). When the answer sheet contains lots of incorrect points, the student will re-answer the exam based on both the standard answer and the judgment from external evaluating system (see the bottom part in Fig. 2c). After that, the losing points is identified by similar comparison. Repeating these steps, the student's ability improves as the score hitting and point losing become clear. It should be noted that the initial "answer sheet" may not necessarily be completed by the efforts of the student. That is if the student's ability is not enough, learning from others' answers is encouraged, especially from ``good students''. If the student is capable, self-studying is a better strategy than the use of the student's answers. 



Specifically, for the reasoning output, NLFT evaluates the conditional probability variations of each token under different prompt conditions to allocate the saliency token. On this basis, the model refines the loss function based on the saliency level of each token, thereby enabling more efficient fine-tuning of the model. An overview of our algorithm is shown in fig.\ref{fig:SFT&ReFT&NLFT}.




In summary, the contributions of this paper are as follows:

\begin{enumerate}
    \item This paper introduces NLFT, a \textbf{token-level natural language} fine-tuning algorithm. In contrast to previous response-level fine-tuning methods that convert natural language into scalar rewards, our approach directly leverages natural language and conducts token-level annotation. This method significantly improves the information utilization efficiency within datasets, thereby reducing the data requirements for NLFT. As a result, by the use of only tens to thousands of data entries, NLFT can achieve domain-specific fine-tuning of LLMs. 

    \item In this paper, we propose a novel Minimal data fine-tuning that eliminates the need for a warm-up phase, which is required by ReFT process.  Furthermore, NLFT achieves an accuracy rate that \textbf{significantly surpasses SFT} with a small number of data entries. Besides, due to the efficient algorithm design in both GPU memory usage and training time, our method offers a substantial advantage in time and memory utilization compared to other methods (such as ReFT).
    \item NLFT exhibits the extraordinary ability to solve the intrinsic pitfall of overfitting phenomenon associated with small-sample data~\cite{9023664}.  Moreover, as a token-level fine-tuning approach, NLFT possesses an outstanding interpretability compared to response-level fine-tuning. NLFT achieves a 64.29\% accuracy with only 50 training data samples, exhibiting a faster improvement rate compared to SFT and outperforming SFT by 219\%, as shown in Fig.~\ref{fig:fig01}.


 
\end{enumerate}


\section{Related Work}
\subsection{Natural Language Learning}

In the past few years, much of the research in the area of  LLM fine-tuning has focused on scalar rewards, which are less efficient than directly using explicit natural language feedback. This is because scalar reward-based methods provide only an indirect understanding of semantic information, which can be suboptimal. In contrast, natural language feedback enables the expression with more delicate and complex preferences. For example, Contrastive Unlikelihood Training (CUT)~\cite{xu2023reasons} uses negative judgments to align values at the token level. Building on this idea, Natural Language Reinforcement Learning (NLRL)~\cite{nlrl} redefines RL principles within a natural language representation space, further demonstrating how natural language can facilitate both efficient policy optimization and improved interpretability.

\subsection{Token-level LLMs Fine-Tuning}
Response-level fine-tuning has played a significant role in pretraining LLMs, but it often faces challenges in terms of training difficulty and stability. In contrast, token-level fine-tuning has emerged as a promising alternative. Recently, several notable works have been proposed to improve token-level fine-tuning. For instance,~\cite{rafailov2024r} extends DPO to a token-level Markov Decision Process (MDP), which enhances alignment with the autoregressive structure of LLMs and optimizes credit assignment. Additionally,~\cite{zhong2024dpo} introduces Reinforced Token Optimization (RTO), which combines token-level rewards with DPO and Proximal Policy Optimization (PPO) to improve policy learning efficiency significantly. These methods demonstrate the substantial potential of token-level fine-tuning in improving model performance, particularly in complex tasks where precision and consistency are crucial.

\section{Method}



\begin{figure}[h]
    \centering
    \includegraphics[width=1\linewidth]{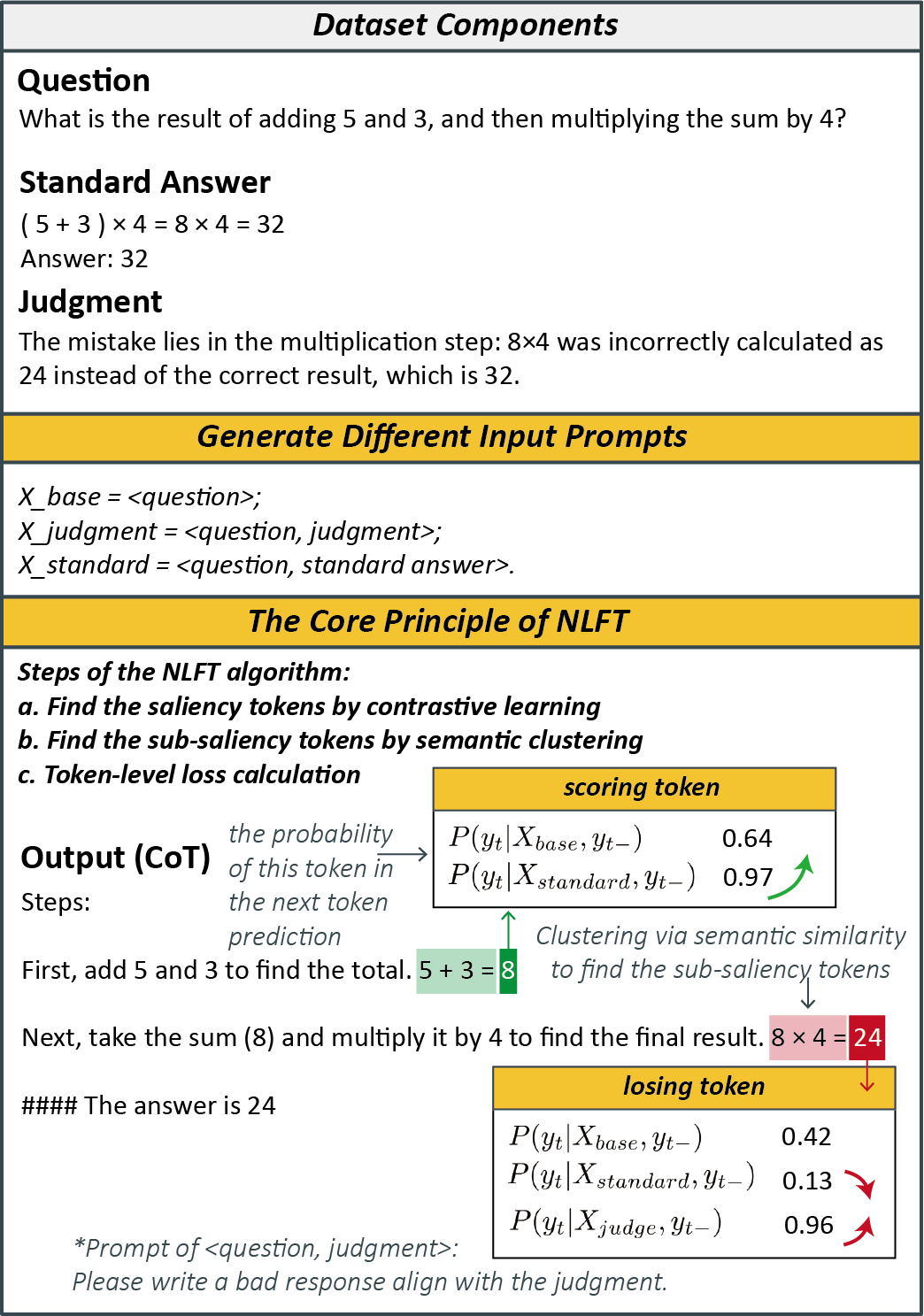}
    \caption{An example of the training process of NLFT, which takes question, standard answer, and judgment as inputs and generates different Input Prompts. Then, under different prompts, the algorithm compares different conditional probabilities to allocate the saliency token and assign scale values.}
    \label{fig:NLFT}
\end{figure}

In this section, we will provide a comprehensive overview of the token-based fine-tuning algorithm NLFT, a novel fine-tuning algorithm that concentrates on natural language CoT and its result. Assuming the input of LLM is $X$ and the reasoning outcomes from CoT is $Y=\{y_1, y_2, ..., y_n\}$, by conducting different input prompt $X$, we can obtain the conditional probabilities of each token within the same CoT output under different input conditions. After that, by comparing these conditional probabilities, the saliency tokens are allocated and we perform token-level loss calculation, thereby achieving fine-grained tuning of the LLM. In short, the NLFT algorithm yields significantly superior results with time and space complexity compared to SFT. The detailed process of NLFT is shown in Algorithm \ref{algNLFT}: Natural Language Fine-Tuning. A more intuitive illustration can be observed in Fig. \ref{fig:NLFT}.


\begin{algorithm}
\caption{Natural Language Fine-Tuning}
\label{algNLFT}
\begin{algorithmic}[1]
\STATE \raggedright\textbf{Input:} $X_{base}\!=\!<\!question\!>$, $X_{judge}\!=<\!question, judgment\!>$, $X_{standard}=\!<question, standard~answer\!>$, a CoT reasoning output $Y=\{y_1, y_2, ..., y_n\}$
\STATE \textbf{Output:} Fine-tuned model
\IF{$Y$ is correct}
\FOR{$t = 1$ to $n$}
    \STATE Collect $P(y_t|X_{base}, y_{t-})$, $P(y_t|X_{standard}, y_{t-})$
    \STATE Calculate $S(y_t)$
\ENDFOR
\ELSE
\FOR{$t = 1$ to $n$}
    \STATE Collect $P(y_t|X_{base}, y_{t-})$, $P(y_t|X_{judge}, y_{t-})$, $P(y_t|X_{standard}, y_{t-})$
    \STATE Calculate $S(y_t)$
\ENDFOR
\ENDIF
\STATE $L=\frac{1}{N}\sum S(y_t) \times log P(y_t |X_{base}, y_{t-})$
\STATE Fine-tune model
\RETURN Fine-tuned model

\end{algorithmic}
\end{algorithm}


\subsection{Preliminary Considerations}


Recently, an interesting work on AI alignment via linguistic feedback, Contrastive Unlikelihood Training(CUT)~\cite{xu2023reasons} employs contrastive learning to fine-tune LLMs to modify erroneous output based on human negative judgments. CUT reveals that, in contrast to the error-free tokens, erroneous tokens experience significant variations in conditional probabilities under the two distinct input conditions of presence and absence of judgment. Building upon this insight, we formulate the following hypotheses through extensive experimental observations: Denoted different prompt inputs as $X$, the CoT reasoning results as $Y=\{y_1, y_2, ..., y_n\}$, and the conditional probability of each token under different input as $P(y_t|X, y_{t-})$, we assume that:

\begin{enumerate}
    \item When the output $Y$ aligns with expectations (that is, the output is correct), the conditional probability $P(y_t|X, y_{t-})$ of key scoring token (Saliency Token) $y_t$ is significantly higher with prompt input $X_{standard}=\!<question, standard \quad answer\!>$ compared to $X_{base}=\!<question\!>$.
    \item When the output $Y$ falls to meet expectation (that is, the output is incorrect), the conditional probability $P(y_t|X, y_{t-})$ of key scoring token (Saliency Token) $y_t$ exhibit substantial variations under three prompt inputs: $X_{base}=\!<question\!>$, $X_{judge}=\!<question, judgment\!>$, and $X_{standard}= \!<question, standard \quad answer\!>$.
\end{enumerate}
Based on these assumptions, we introduce the NLFT algorithm. Firstly, depending on whether the output meets expectations, the conditional probabilities of each token are obtained under different prompt inputs. After that, Saliency Tokens are located through contrastive learning, and the phrases containing these tokens are located using a semantic similarity cluster. Finally, a token-level loss function is constructed to achieve fine-grained fine-tuning.

\subsection{Token-level Conditional Probability Analysis}

Formally, the process of generating results through the natural language CoT can be decomposed into a sequence of next-token prediction actions. To be specific, take the CoT reasoning result as $Y=\{y_1, y_2, ..., y_n\}$, where each $\!<y_t\!>$ is a token inferred based on the given input $X$ and the previous output $y_{t-}=\{y_1, y_2, ..., y_{t-1}\}$. Besides, each $y_t$ has its own conditional probability function $P(y_t |X, y_{t-})$.  As previously stated, the conditional probabilities of salient tokens undergo significant changes with varying inputs $X$. Building on this observation, we categorize the reasoning results $Y$ into correct and incorrect outcomes, then selectively perform conditional probability lookup and collection for each category.

\textbf{Correct Output.} When the CoT reasoning output $Y$ is correct, the input $X$ is divided into two categories: $X_{base}=\!<question\!>$ and $X_{standard}=\!<question, standard \quad answer\!>$. Then, for each token $y_t$ of output $Y$, we will have two conditional probability $P(y_t |X_{base}, y_{t-})$ and $P(y_t |X_{standard}, y_{t-})$. In following section, we will allocate the saliency token based on these conditional probabilities.


\textbf{Incorrect Output.} When the reasoning output $Y$ is incorrect, the input $X$ is divided into three categories: $X_{base}=\!<question\!>$, $X_{judge}=\!<question, judgement\!>$, and $X_{standard}=\!<question, standard~answer\!>$. Then, for each token $y_t$ of output $Y$, we will have three conditional probability$P(y_t |X_{base}, y_{t-})$, $P(y_t |X_{judge}, y_{t-})$, and $P(y_t |X_{standard}, y_{t-})$. In the following section, we will allocate the saliency token based on these conditional probabilities.

\subsection{Probability-driven Saliency Token Allocation}

After obtaining the conditional probabilities, we proceed to classify them and allocate the saliency tokens. Similarly, the allocation strategies are divided into two categories: correct outcomes and incorrect outcomes. 

\textbf{Correct Output.} When the CoT reasoning output $Y$ is correct, we set the threshold conditional probability $p_0^{correct}$. When the conditional probability $P(y_t |X_{standard}, y_{t-}) > p_0^{correct}$, it is considered that the token is more likely to be adopted under $X_{standard}$ condition than the threshold conditional probability. This implies a greater correlation between the token and the $X_{standard}$ condition, hence it is allocated as a saliency token. After that, we perform semantic clustering around these saliency tokens to identify their associated phrase and designate them as sub-saliency tokens. Finanlly, the remaining token are categorized as irrelevant tokens.

\textbf{Incorrect Output.} For the incorrect CoT reasoning output $Y$, we cannot simply set a threshold conditional probability, as the conditional probabilities for all tokens are generally lower in such instances. Therefore, we calculate the following two ratios,
\begin{equation}
    r_1 = \frac{P(y_t |X_{judge}, y_{t-})}{P(y_t |X_{base}, y_{t-})},
    \label{r1}
\end{equation}
\begin{equation}
    r_2 = \frac{P(y_t |X_{judge}, y_{t-})}{P(y_t |X_{standard}, y_{t-})}.
    \label{r2}
\end{equation}
If a token is a saliency token, then its conditional probability under $X_{judge}$ condition should be much higher than that under $X_{base}$ and $X_{standard}$. Consequently, its corresponding ratios $r_1$ and $r_2$ will also be higher. Therefore, if a token has both its corresponding ratios $r_1$ and $r_2$ exceeding a preset value $r_0$, and its own conditional probability surpasses the threshold $p_0^{incorrect}$, the token will be allocated as a saliency token. After that, we also perform semantic clustering around these saliency tokens to identify the associated phrase. Given that these tokens are close to the saliency token under incorrect situations, we directly categorize them as irrelevant tokens and assign a scale of zero to them in subsequent contrastive learning processes.

\subsection{Token-level Loss Calculation}
In this section, we will assign scale values based on the previously allocated saliency tokens and proceed with contrastive learning. Similarly, the scale strategies are divided into two categories: correct outcomes and incorrect outcomes. 

%
\textbf{Correct Output.} As we shown before, in the correct CoT reasoning output, the tokens are classified into three kinds: saliency tokens, sub-saliency tokens, and irrelevant tokens. For each token, we have scales,

\begin{equation}
\label{scalecor}
S(y_t) =
\begin{cases}
\displaystyle
1 + \left(\frac{P(y_t \mid X_{standard}, y_{t-}) - p_0^{correct}}{1 - p_0^{correct}}\right)^{c_1}, \\
\hfill \text{if } y_t \in Y_{saliency}, \\[6pt]
\displaystyle
\left(\frac{P(y_t \mid X_{standard}, y_{t-})}{p_0^{correct}}\right)^{c_2}, \\
\hfill \text{if } y_t \in Y_{sub-saliency}, \\[6pt]
\displaystyle
\left(\frac{P(y_t \mid X_{standard}, y_{t-})}{p_0^{correct}}\right)^{c_3}, \\
\hfill \text{if } y_t \in Y_{irrelevant}.
\end{cases}
\end{equation}

where $c_1$, $c_2$, and $c_3$ are hyper-parameters and $c_2< c_3$. It can be observed that under such configuration, all three scales increase as the conditional probability $P(y_t |X_{standard})$ grows. Meanwhile, the scales of the saliency tokens are the largest and always exceed 1, while the scales of the sub-saliency tokens are consistently greater than that of the irrelevant tokens.

\textbf{Incorrect Output.} In the incorrect CoT reasoning output, the tokens are classified into two kinds: saliency tokens and irrelevant tokens. For the irrelevant tokens, we will set the scales to $0$, as within incorrect Output, we only wish to consider the saliency tokens. For the saliency tokens, we have scales,
\begin{equation}
    S(y_t)=\frac{2}{1+e^{-(r_1-r_0)}}
\end{equation}

The scales are larger than $0$ and smaller than $1$, and these scales also increase as the conditional probability $P(y_t |X_{judge})$ grows. 

After obtaining the scales of each token, we get our final loss function,

\begin{align}
L &= \frac{1}{N} \left( \sum_{y_t \in Y_{correct}} S(y_t) \times \log P(y_t |X_{base}, y_{t-}) \right. \nonumber \\
  &\quad \left. + \sum_{y_t \in Y_{incorrect}} S(y_t) \times (1 - \log P(y_t |X_{base}, y_{t-})) \right)
\end{align}

\section{Experiments}

\subsection{Dataset}

We conduct experiments on the Mathematics problem dataset GSM8K~\cite{cobbe2021GSM8K}. It offers mathematics problems in natural language form, standard solution processes, and numerical standard answers. The training set of GSM8K contains 7473 entries, while the test set contains 1319 entries. We employed data prompting and CoT prompting to obtain the analysis and result (that is, the reasoning output). 



\textbf{Learning from teaching: }When the accuracy of the base model to be fine-tuned is low, we choose to use other models with better performance to generate the analysis and result (that is, the reasoning output). We refer to this process as ``teaching". Specifically, We utilized LLAMA3-8B-Instruct~\cite{roziere2023code} to obtain the analysis and result of the case. 

\textbf{Learning from self-study: }When the model has the capacity to generate a certain percentage of correct responses, we proceed to let the trained model learn from its own answers and produce results. We refer to this process as self-study. In this scenario, we directly utilize the trained model to produce the reasoning output.


When the reasoning output is incorrect, we leverage the GPT-4o~\cite{openai2023gpt} to acquire the annotations for judgment. The instruction prompt for judgment is provided in Appendix.~\ref{appendix1:PromptStrategiesJudgment}.



\subsection{Experimental Setup}

We conduct experiments on the LLAMA3-8B base model~\cite{roziere2023code}.


\textbf{Training Subset Setup:} To investigate the learning capacity of NLFT with a small dataset, we randomly shuffle the data in the training set. According to the shuffled index order, we take the first 400, first 800, first 25\%, first 50\%, and 100\% respectively to construct training sets for experimental preparations.

\textbf{Hyper-parameter Settings:} All experiments are carried out on two A800 GPUs, which is four times lower than the requirement demanded by reinforcement learning-based fine-tuning methods such as ReFT. Besides, we select the AdamW optimizer~\cite{loshchilov2019DecoupledWeight} and the cosine learning rate scheduler. The batch size is set to be 4 and the learning rate is set to be $5 \times 10^{-5}$. If a small dataset is utilized, the model is trained for 10 epochs, while with an extensive dataset, training is trained for 3 epochs. For the parameters in NLFT, $p_0^{correct}$ is set to be 0.95, $p_0^{incorrect}$ is set to be 0.01, and $r_0$ is set to be 1.5. For the scale hyper-parameters, $c_1$, $c_2$, and $c_3$ are set to be 5, 0.3, and 0.6. Detailed hyper-parameter configurations are shown in Appendix~\ref{appendix2}.


\textbf{Evaluation: }We utilize the full dataset to assess the accuracy of natural language CoT reasoning Output from fine-tuned LLM. The evaluation process employs the same prompt templates as those used during the Training phase. In our settings, the temperature is set to 0.6 during text generation, and maximum generation length is 512 tokens.


\subsection{Baseline}

We compare our model NLFT with SFT~\cite{ouyang2022training} and ReFT~\cite{luong2024reft} baselines. To ensure a fair comparison, we make sure that the hyperparameter settings for the SFT baseline match those of the NLFT experiments. The details on the hyperparameter settings is shown in Appendix~\ref{appendix2}. Besides, our study concentrates on natural language fine-tuning, hence we only select the corresponding component of ReFT, that is, the CoT-N portion. Although the procedural language (CoT-P) component of ReFT shows better performance on the GSM8K dataset for mathematical reasoning tasks, it significantly deviates from our experimental setup, leading to its omission.




\subsection{Results}

\begin{figure}[h]
    \centering
    \includegraphics[width=1\linewidth]{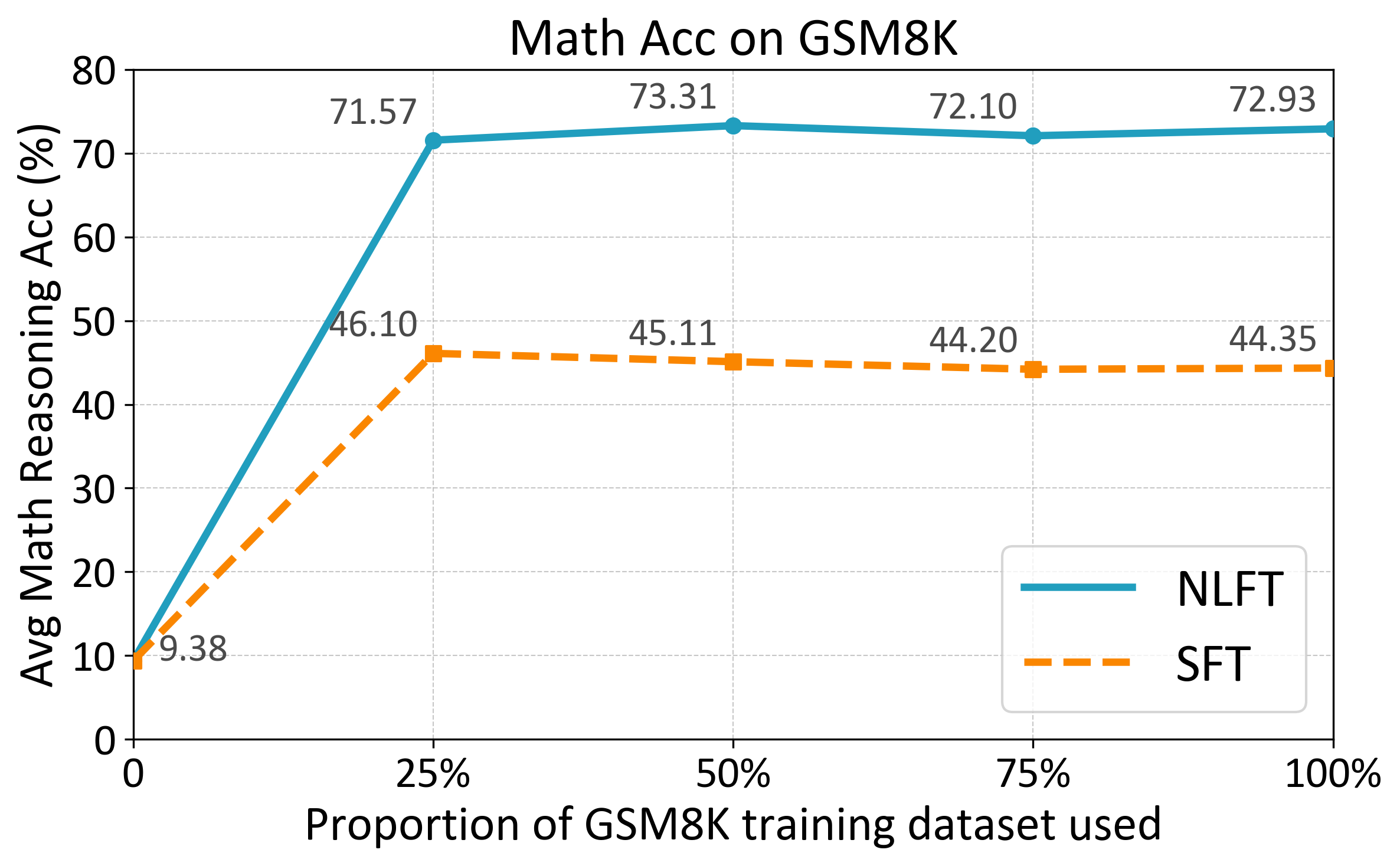}
    \caption{Comparison of accuracy of SFT and NLFT using 25\%, 50\%, 75\%, and 100\% of GSM8K training set, corresponding to 1868, 3737, 5605, and 7473 samples, respectively. At proportion of 0 represents base model before fine-tuning.}
    \label{fig:NLFTPercentageAcc}
\end{figure}

\begin{figure}[h]
    \centering
    \includegraphics[width=1\linewidth]{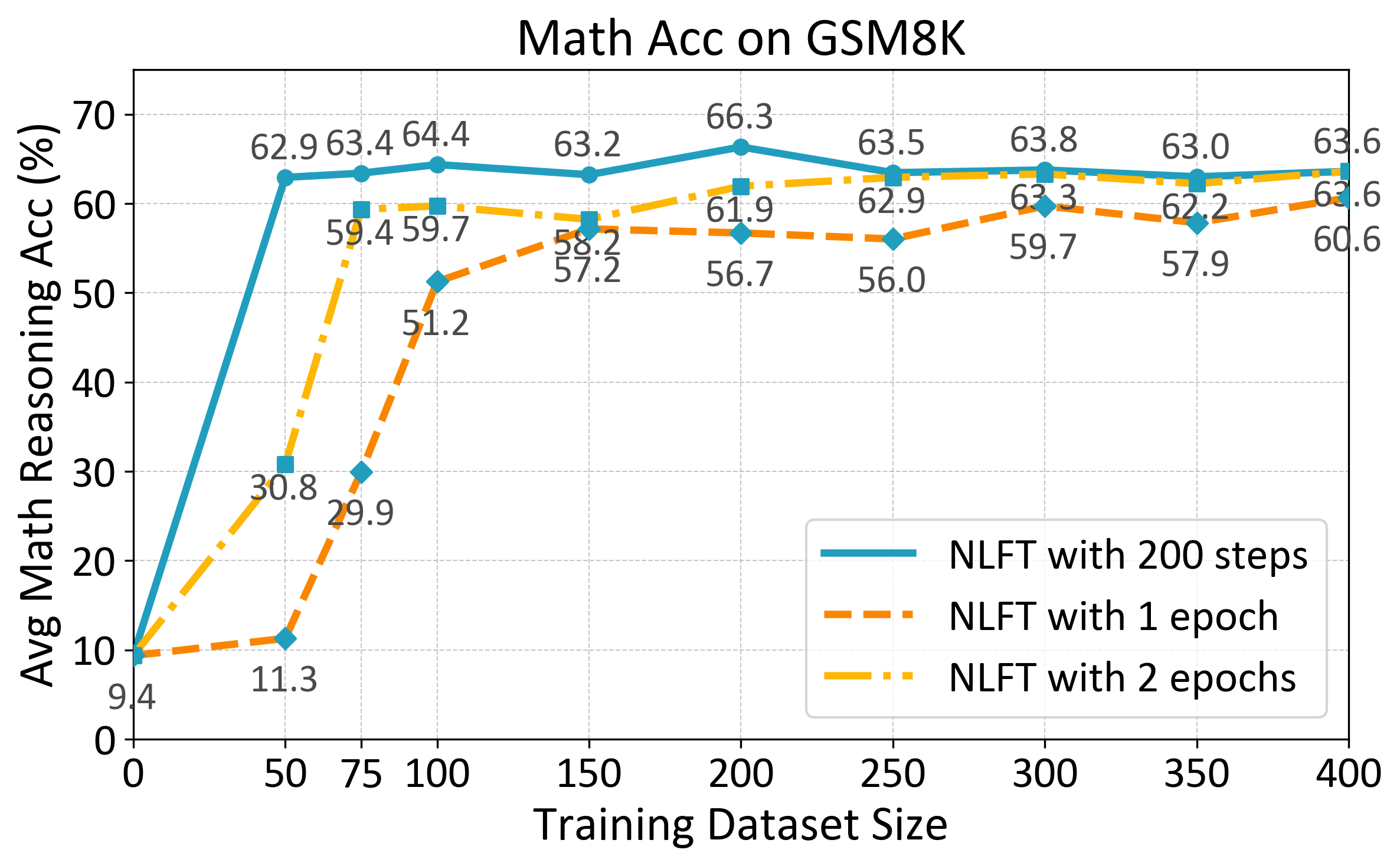}
    \caption{Comparison of accuracy of NLFT using minimal dataset samples of GSM8K as a training set, including NLFT trained with 200 steps, 1 epoch, and 2 epochs, respectively. To better illustrate the increase in accuracy from the data size of 50 to 100, we additionally provide plots for data size of 75, under the settings of 1-epoch and 2-epoch training. }
    \label{fig:NLFT200StepsAcc}
\end{figure}

\begin{figure}[h]
    \centering
    \includegraphics[width=1\linewidth]{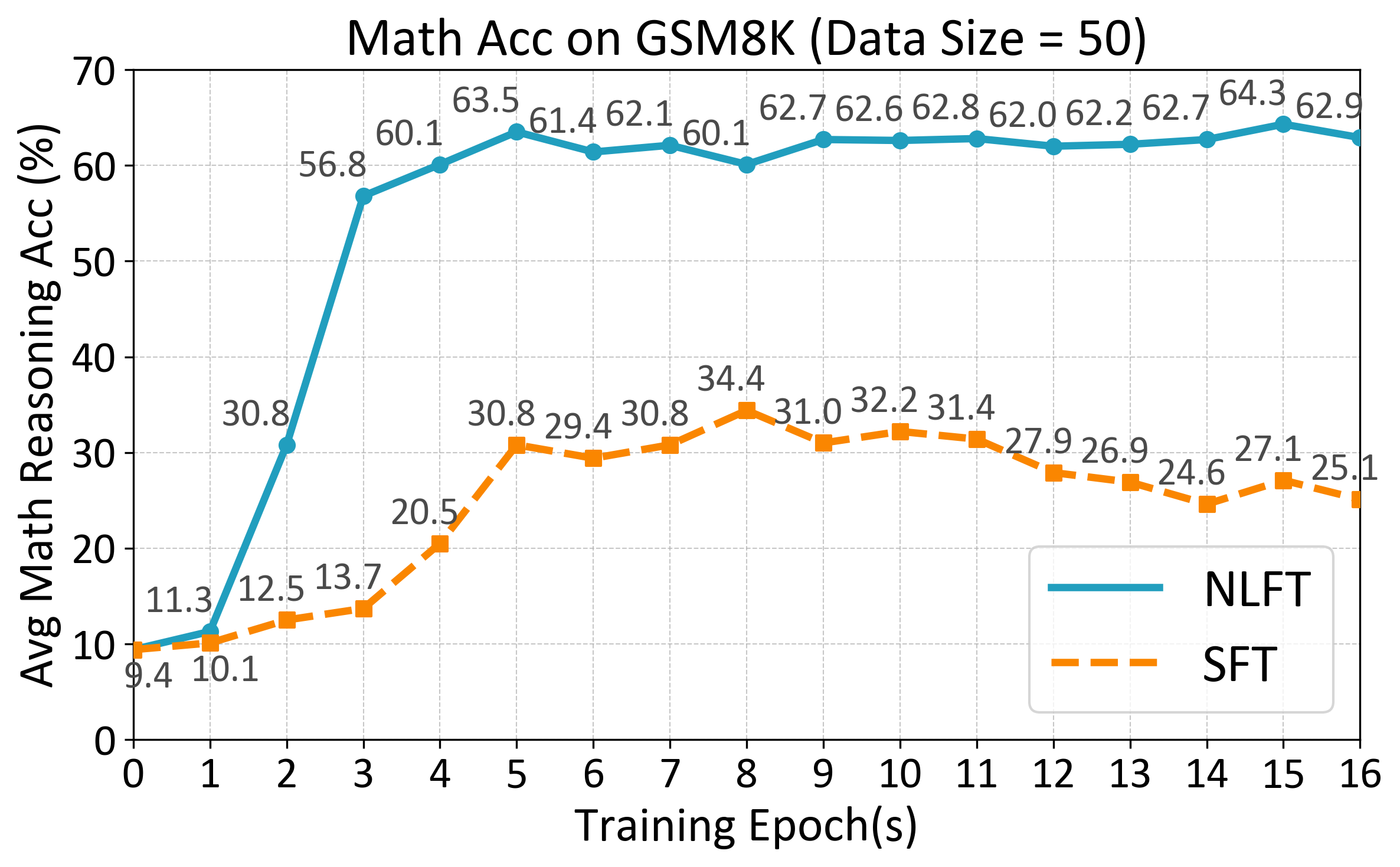}
    \caption{Accuracy comparison between SFT and NLFT trained in minimal dataset samples with data size of 50 from 1 to 16 epochs.}
    \label{fig:NLFT50vsSFTin16Epochs}
\end{figure}

\begin{figure}[h]
    \centering
    \includegraphics[width=1\linewidth]{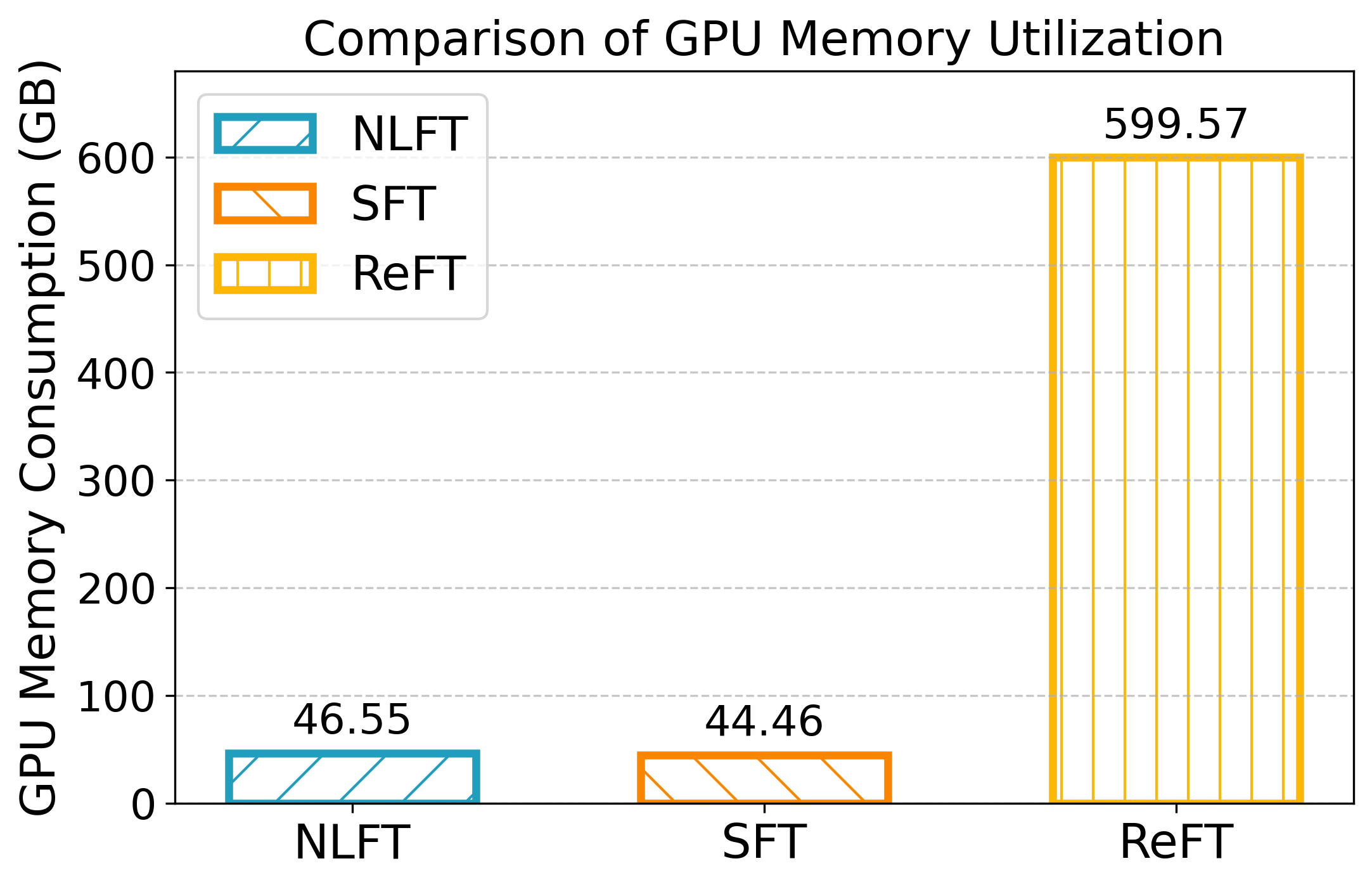}
    \caption{Comparison of GPU memory utilization between SFT, ReFT, and NLFT. The batch size in ReFT is set to 2. SFT and NLFT share the same runtime batch size of 2. NLFT has GPU memory usage as lightweight as SFT, which is more than 10 times lower than ReFT.}
    \label{fig:gpu-memory-utilization}
\end{figure}

\begin{figure}[h]
    \centering
    \includegraphics[width=1\linewidth]{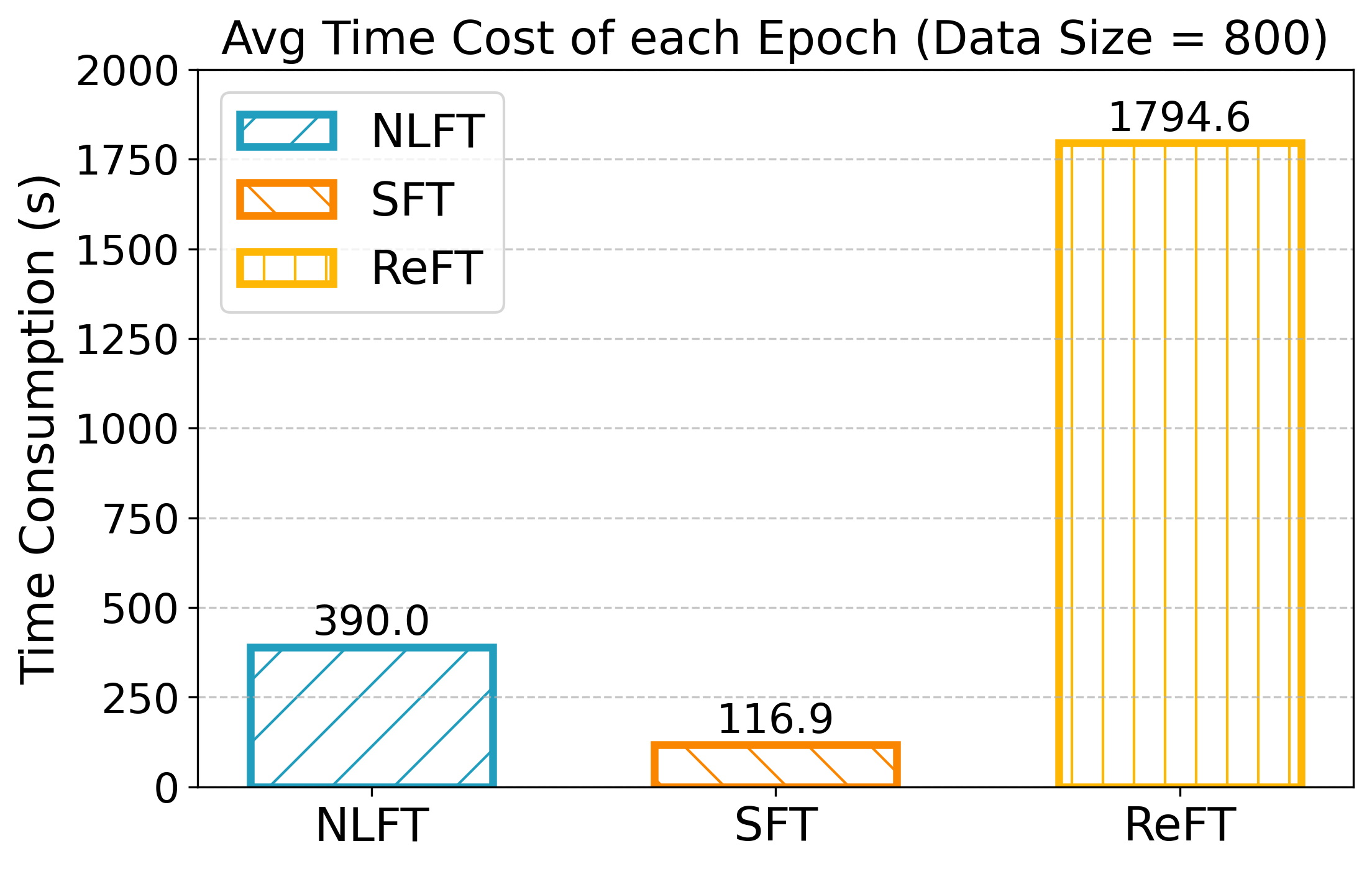}
    \caption{Comparison of per-epoch average time cost of NLFT, SFT, and ReFT. The data size of each experiment is fixed to 800. ReFT has a significantly higher time cost compared to NLFT and SFT. NLFT takes approximately 3 times longer than SFT on average.}
    \label{fig:time-cost}
\end{figure}

\begin{figure*}[h]
    \centering
    \includegraphics[width=1\linewidth]{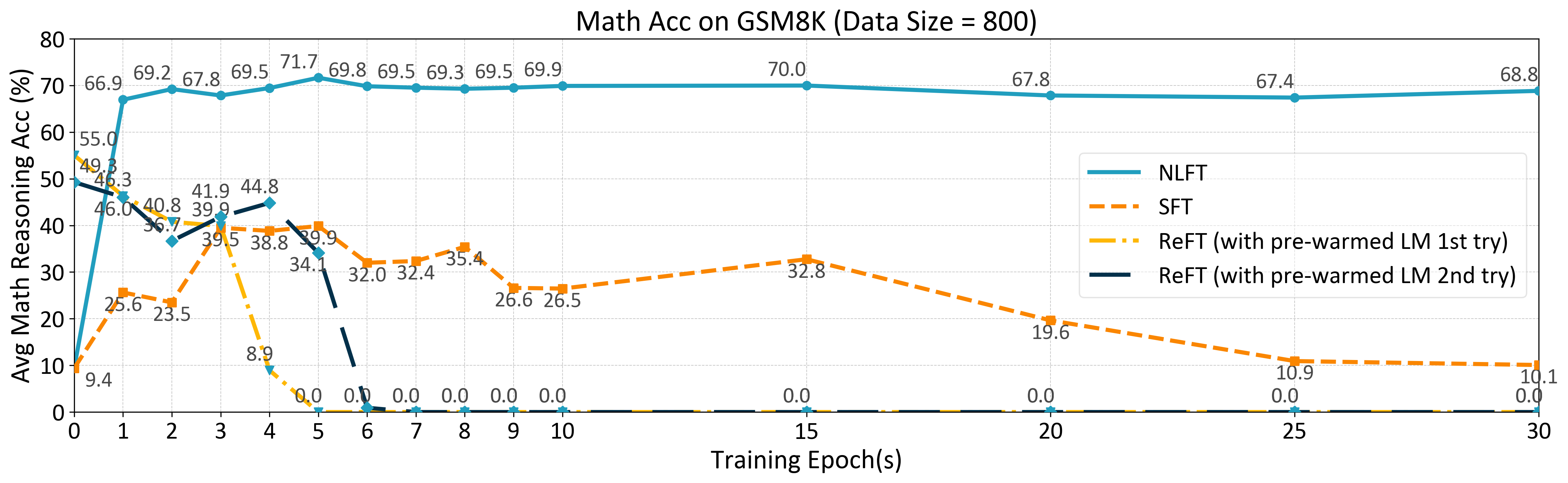}
    \caption{Comparison of accuracy of NLFT, SFT, ReFT with data size of 800. Both experiments of ReFT are pre-warmed using SFT, and the initial accuracy is shown in epoch 0.}
    \label{fig:Data800Epoch30Acc}
\end{figure*}


\textbf{Full dataset experiment: } Fig.~\ref{fig:NLFTPercentageAcc} compares the performance of NLFT and SFT baseline on GSM8K dataset. Starting from the same initial base model at proportion of 0 \footnote{In our experiments, LLAMA3-8B base model failed to provide any reasoning for getting answer, and its generation mostly just repeats the instruction prompt. The accuracy claimed in LLAMA3 official website is achieved by LLAMA3-8B-Instruct under our reproduction.}, we can observe that NLFT consistently achieves higher accuracy over SFT, with NLFT achieving an accuracy rate above 70\% across all four percentage datasets, whereas SFT is between 44\% and 46\%. NLFT outperformed SFT with an accuracy improvement of over 25\%. Furthermore, the change in the proportion of the training dataset has little impact on the accuracy of both NLFT and SFT, which indicates the boundary effect on accuracy improvement by expanding data size will gradually decrease. Therefore, it is reasonable to shift our focus to a smaller size of training dataset especially lower than 25\%, to fill the gaps of accuracy details from 0 to 25\%.


\textbf{Limited-size dataset experiment: }
To investigate the performance of NLFT when using minimal dataset samples as a training set, we adopt data sizes ranging from 50 to 400, separated by 50. In Fig.~\ref{fig:NLFT200StepsAcc}, we conduct experiments with fixed training steps at 200 across different data sizes. We observe that the first plot, utilizing shuffled 50 data entries, has achieved an accuracy rate of 62.93\%, which is close to the last plot with 400 data entries.

To reveal the intermediate states of models when LLM learns from different data entries, we evaluate model accuracies trained after 1 and 2 epochs and plot them as dotted lines. In our settings, the more epochs or steps the model is trained, the closer it converges to the line of NLFT with 200 steps.

Fig.~\ref{fig:NLFT50vsSFTin16Epochs} compares the accuracy of NLFT over 16 continuous epochs. We observe that the accuracy in epoch 1 was 11.30\%. By epoch 2, it sharply increased to 30.8\%, reaching the accuracy of SFT by epoch 5. Subsequently, the model accuracy continued to improve, reaching 60.1\% by epoch 4, after which it remained consistently above 60\%. Meanwhile, SFT started its rapid ascent from epoch 1 to 5, achieving the highest accuracy of 34.4\% at epoch 8, after which it gradually declined. These results indicate that NLFT demonstrates breakthrough learning potential with a limited dataset, which, to the best of our knowledge, is not possessed by fine-tuning algorithms such as SFT.



\textbf{Algorithm comparison:} To validate the performance of different fine-tuning algorithms under identical data scales and training conditions, Fig.~\ref{fig:Data800Epoch30Acc} conducted experiments using a random subset of the first 800 data points, training and testing the accuracy of NLFT, SFT, and ReFT after 10 epochs. The hyperparameters for NLFT and SFT were configured consistently, while those for ReFT were set according to the original paper. Following~\cite{luong2024reft}, ReFT trains on the basis of SFT warm-up with 2 epochs. In the first test, we observed that ReFT accuracy persistently declined over the first 4 epochs, and dropped to zero by epoch 5 \footnote{We reproduced ReFT, and experiments show that under full dataset of GSM8K, ReFT can improve accuracy on top of SFT-warmup models, just as in the original paper. However, here we focus on experiments of minimal data fine-tuning, and in order to make consistency with NLFT and SFT experiments, we adopted training results with 800 data points instead of full dataset.}. To ensure the experimental result is correct, we conducted a second experiment with ReFT, maintaining a data scale of 800 instances, and observed a similar decline to zero accuracy by the 6 epoch. Upon reviewing the output, we discovered that the model regraded to repeating the instruction. This degradation is attributed to the instability of ReFT (and other reinforcement learning-based fine-tuning algorithms), which have significant requirements for data quantity. While ReFT can learn effectively with the full dataset, it often reverts to a state of non-learning when data quantity is insufficient. Under 800 data samples, SFT is capable of learning effectively, reaching an accuracy of 39.88\% at epoch 5, after which the accuracy had a decline of over 10\%. In contrast, NLFT achieved an accuracy of 71.65\%, with no significant drop in accuracy following the epoch of peak performance. This demonstrates the universality of NLFT with respect to data quantity and the efficiency of its training outcomes.


\textbf{Time Cost Analysis:}
Under 800 data samples, we conducted a comparative analysis of the training time required for NLFT, SFT, and ReFT across 10 epochs. Since ReFT algorithm cannot be executed in two-GPU configuration, we recorded time consumption of each algorithm under an eight-GPU configuration for fair comparison. As shown in Fig.~\ref{fig:time-cost}, ReFT required an average of 30 minutes per epoch, whereas NLFT, due to the use of more GPUs, saw a significant reduction in training time compared to 26.1 minutes the two-GPU setup, averaging 6.5 minutes.

It is worth noticing that, the time consumption ratio between NLFT and SFT is around 3. NLFT involves three times the forward inference processes compared to SFT, hence its time complexity constant is at least 3. Despite the increased constant term, NLFT still qualifies as a lightweight fine-tuning algorithm with linear time complexity.


\textbf{Memory Use Analysis:} Fig.~\ref{fig:gpu-memory-utilization} illustrates the runtime GPU memory usage of each fine-tuning algorithm. With a two-GPU configuration, SFT averages a total memory usage of 44.55 GB, while NLFT averages 46.87 GB. NLFT's memory usage is only 5.2\% higher than SFT's, which still falls within the category of lightweight fine-tuning algorithms. In contrast, ReFT requires an average of 599.57 GB of total memory, which is not in the same order of magnitude as NLFT. Regarding hardware configuration and memory usage, NLFT not only matches SFT's requirements but also significantly outperforms reinforcement learning-based fine-tuning algorithms like ReFT, offering a unique advantage in terms of hardware resource demands.



\section{Analysis}
\subsection{Stability Analysis}

Compared with SFT and ReFT, the NLFT shows a robust capacity to mitigate overfitting in minimal data scenarios. Fig.~\ref{fig:NLFT50vsSFTin16Epochs} shows the performance of SFT and NLFT with a minimal data dataset of 50 samples across 1-16 epochs. It can be observed that SFT exhibits pronounced overfitting, while NLFT maintains a consistent level of accuracy. Fig. \ref{fig:Data800Epoch30Acc} shows the performance of ReFT, SFT, and NLFT with a dataset of 800 samples. It can be observed that SFT exhibits slow improvement in the initial epochs and reaches near-optimal performance after 5 epochs. However, as the number of epochs increases, the risk of overfitting grows, leading to a significant overfitting and a sharp decline in accuracy. In contrast, NLFT exhibits a stable accuracy of approximately 70\%, reflecting its robust stability. We suppose this is because the algorithm shows more attention to the saliency token, thereby focusing on the most critical problem-solving pathway and improving training stability. This approach is similar to causal-inspired stable learning in computer vision~\cite{zhang2021deep}, which effectively filters out irrelevant features and uses only truly relevant ones for prediction, resulting in more stable performance in wild, non-stationary environments.

\subsection{Algorithm Complexity Analysis}

The time complexity and space complexity of our algorithm are both $O(n)$, which means that as the input size $n$ grows, the required time and space resources grow linearly. This linear growth indicates that our algorithm is efficient and resource consumption is controllable when dealing with large-scale data. In contrast, the ReFT algorithm uses the PPO algorithm~\cite{schulman2017proximal} for optimization, whose time complexity is $O(TC+NBP)$, which is proportional to the product of the number of samples, the neural network’s forward and backward propagation complexity, and the number of update iterations~\cite{luong2024reft}. Due to its higher time and space complexity, ReFT may suffer from decreasing efficiency and increasing resource consumption when processing large data. Therefore, in applications involving large-scale data processing, our algorithm demonstrates a significant performance advantage over the ReFT.

\section{Discussion}
\begin{figure}[h]
    \centering
    \includegraphics[width=1\linewidth]{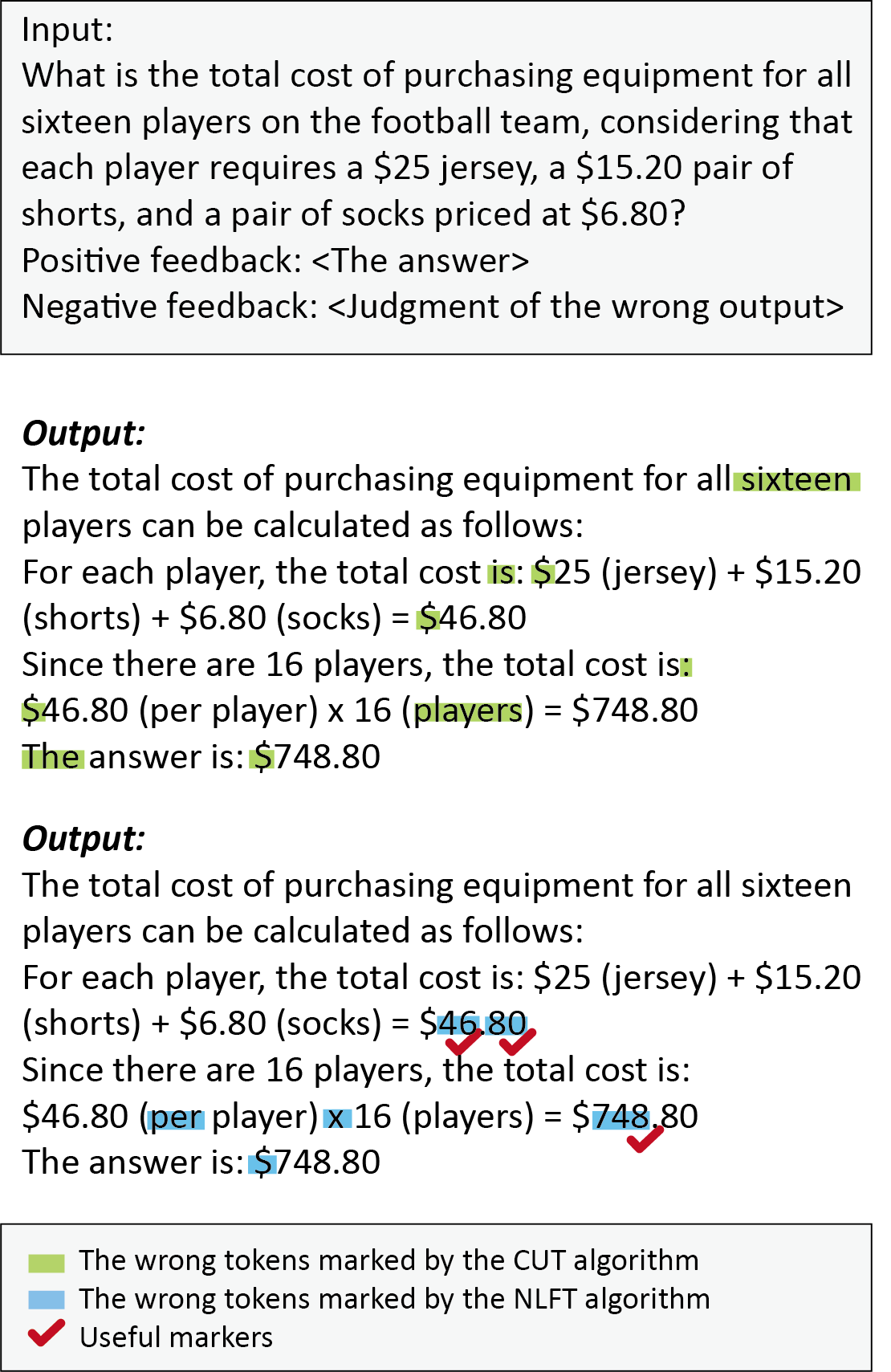}
    \caption{Token-level selection comparison between CUT and NLFT on a single instance from the GSM8K dataset. Green tokens are marked by CUT, blue tokens by NLFT with ratio values, red tokens represent incorrect selections, and red circles indicate correctly marked tokens.}
    \label{fig:NLFTvsCUT}
\end{figure}
\subsection{Learning Efficiency from Incorrect Samples}
Our experiments indicate that increasing the proportion of incorrect samples in the training data leads to a decrease in learning efficiency. To fine-grained investigate the contribution of incorrect samples to model fine-tuning, we attempt to visualize saliency tokens at the token level to reflect the intermediate process of LLM training. In Fig. \ref{fig:NLFTvsCUT}, we compare NLFT with another token-level LLM fine-tuning algorithm, CUT~\cite{xu2023reasons}, and mark the recognized incorrect token of NLFT and CUT. The results demonstrate that compared to the CUT algorithm, NLFT can more accurately identify incorrect tokens in answers. Additionally, we implement a filtering strategy for entirely incorrect samples. When the proportion of erroneous tokens exceeds a certain threshold, NLFT will exclude that training data.

However, after such processing, the efficiency of learning from incorrect samples still remains significantly lower compared to learning from correct samples. We analyze that, on one hand, the value of acquiring new knowledge may outweigh the restrain of incorrect tokens; on the other hand, after multiple rounds of training, the model may exhibit a phenomenon of "logical coherence", it avoids marking incorrect tokens, thereby reducing the MLE loss. Regarding how to enhance the learning efficiency from incorrect samples, we will continue to explore this in our subsequent research.




\subsection{Model Generalization}
In our previous research, we applied the simplified version of the NLFT to human-computer collaborative tasks and achieved significant performance improvement. Theoretically, NLFT is applicable to scenarios where outputs can be generated through CoT and labeled data is available, such as coding, medical diagnosis, natural language inference, and complex question-answering systems. By comparing the generated output with the labels, it is possible to annotate the saliency tokens, thereby applying NLFT for token-level fine-tuning. In our future work, we will explore the application of NLFT to broader fields and refine the NLFT algorithm based on the characteristics of each task.




\section{Conclusion}

In this paper, we propose a novel natural language minimal data fine-tuning algorithm NLFT. The algorithm compares the conditional probabilities of various natural language tokens under different prompts, utilizing natural language as a supervisory signal to identify saliency tokens and assign them scaling values.  Experimental results demonstrate that our algorithm, compared to previous ones, has lower time complexity and better performance. Under the GSM8K dataset evaluation, only random 50 training data allows NLFT to achieve over 60\% accuracy, and performance of NLFT is stably increased by 25\% compared to SFT. In contrast to RL-based fine-tuning algorithm like ReFT, NLFT saves huge time and space complexity, enabling broader imagination for lightweight fine-tuning and applications.



\bibliographystyle{named}
\bibliography{nlft}


\appendix

\section{Prompt Strategies}
\subsection{Prompt Strategy of Judgment on the GSM8K Dataset}
\label{appendix1:PromptStrategiesJudgment}
In this section, we present a prompt engineering strategy for generating the judgment of incorrect samples.
\begin{tcolorbox}[colframe=black!20!white, colback=white, coltitle=black, title=\textbf{Instruction}]

Suppose you are a math expert and you are presented with a math problem, a student's response, and the correct answer. 

Please first check whether the student's response is correct. Including checking the solution process and whether the answer is correct.

If the student's response is correct, please directly output: \#\#\# The response is correct. \#\#\#

If the response is wrong, please analyze why the solution is wrong.

\end{tcolorbox}
\subsection{Prompt Strategy of Math Reasoning on the GSM8K Dataset}
\label{appendix1:PromptStrategies}
In this section we present a prompt engineering strategy for testing model performance.

\begin{tcolorbox}[colframe=black!20!white, colback=white, coltitle=black, title=\textbf{Instruction}]


Suppose you are a math expert. The following describes a math problem. Please read it carefully and solve it STEP BY STEP!!!, and give the correct answer.

Please ensure that your output strictly follows the following format requirements: 

\{Your analysis\}\textbackslash n\#\#\#\# \{The answer number\}

Your analysis should be very detailed. And make sure the string "\#\#\#\#" only appears once following the answer number in the end.

\textbf{For example:}

\textless Output Example\textgreater 

Natalia sold 48 $/$ 2 = \textless \textless 48$/$2=24\textgreater \textgreater 24 clips in May.

Natalia sold 48+24 = \textless \textless 48+24=72\textgreater \textgreater 72 clips altogether in April and May.

\#\#\#\# 72

\textless /Output Example\textgreater 

\textless Output Example\textgreater 

The number of truck stamps is 11 + 9 = \textless \textless 11+9=20\textgreater \textgreater 20.

The number of rose stamps is 20 - 13 = \textless \textless 20-13=7\textgreater \textgreater 7.

Bella bought 11 + 20 + 7 = \textless \textless 11+20+7=38\textgreater \textgreater 38 stamps in all.

\#\#\#\# 38

\textless /Output Example\textgreater 

\textless Output Example\textgreater 

Lisa earned \$60 * 1$/$2 = \$\textless \textless 60*1$/$2=30\textgreater \textgreater 30.

Tommy earned \$30 * 1$/$2 = \$\textless \textless 30*1$/$2=15\textgreater \textgreater 15.

Lisa earned \$30 - \$15 = \$\textless \textless 30-15=15\textgreater \textgreater 15 more than Tommy.

\#\#\#\# 15

\textless /Output Example\textgreater 

\textless Output Example\textgreater 

He needs to save up \$400 because 4 x 100 = \textless \textless 4*100=400\textgreater \textgreater 400

He has 8 months to earn this money because 12 - 4 = \textless \textless 12-4=8\textgreater \textgreater 8

He needs to earn \$50 a month because 400 $/$ 8 = \textless \textless 400$/$8=50\textgreater \textgreater 50

He needs to do 5 tasks a month because 50 $/$ 10 = \textless \textless 50$/$10=5\textgreater \textgreater 5

\#\#\#\# 5

\textless /Output Example\textgreater 

\textless Output Example\textgreater 

15 coins collected in hour one

35 coins collected in hour two

35 coins collected in hour three

50 coins collected in hour four

Before giving her coworker some coins there were 15+35+35+50=\textless \textless 15+35+35+50=135\textgreater \textgreater 135 coins

The number of coins after giving 15 to her coworker is 135-15=\textless \textless 135-15=120\textgreater \textgreater 120

\#\#\#\# 120

\textless /Output Example\textgreater 

\end{tcolorbox}

\section{Detailed Hyperparameter Settings}
\label{appendix2}

\textbf{NLFT:} NLFT is trained using LoRA~\cite{hu2021lora}, where the parameters $r$, $\alpha$ and dropout are set to 16, 16, and 0.05, respectively. The learning rate is set to  $5\times 10^{-5}$. For most NLFT experiments, the maximum number of epochs is set to 10, unless the setting in Fig.~\ref{fig:Data800Epoch30Acc} specifies 30.

\textbf{SFT:} Our SFT implementation employs SFTTrainer in trl~\cite{vonwerra2022trl}. To ensure that the SFT code configuration is largely consistent with NLFT configuration, we have essentially adopted most of the parameter settings of NLFT.

\textbf{ReFT:} Following~\cite{luong2024reft}, before ReFT algorithm, we perform SFT warmup for 2 epochs with the learning rate of $1\times 10^{-5}$ on GSM8K dataset. When performing SFT warmup, the batch size is set to 48, and the maximum input length is set to 512. After warmup phase is finished, we perform ReFT algorithm. The maximum input length is set to 300, and the maximum length of model generation is set to 700. The batch size is set to 16 to avoid crash during training. The number of updates per RL step is set to 2. The learning rate is set to $3 \times 10^{-7}$.

\end{document}